%% file: main.tex
\documentclass[sigconf]{acmart}

\setcopyright{none}
\renewcommand\footnotetextcopyrightpermission[1]{}

\AtBeginDocument{%
  }

\input{section/math_commands}
\input{section/preamble}

\newcommand{\name}{EiFormer}

\begin{document}

\title{Towards Efficient Large Scale Spatial-Temporal Time Series Forecasting via Improved Inverted Transformers}
\input{section/author}

\renewcommand{\shortauthors}{Sun et al.}

\input{section/abstract}




\maketitle

\input{section/introduction}
\input{section/related}
\input{section/background}
\input{section/method}
\input{section/experiment}
\input{section/conclusion}

\balance

\bibliographystyle{ACM-Reference-Format}
\bibliography{reference}

\end{document}

%% file: section/preamble.tex
\usepackage{graphicx}
\usepackage{stfloats}
\usepackage{wrapfig}
\usepackage{lipsum}  
\usepackage[capitalise]{cleveref}

\newlength\savewidth
\newcommand\shline{\noalign{\global\savewidth\arrayrulewidth
                            \global\arrayrulewidth 1.5pt}%
                   \hline
                   \noalign{\global\arrayrulewidth\savewidth}
                   }

\usepackage{xcolor}

\usepackage{algorithm}
\usepackage[noend]{algpseudocode}

\algrenewcommand\alglinenumber[1]{#1}

\usepackage{multirow}

\usepackage{enumitem}
\setlist[itemize]{leftmargin=*, topsep=.0em, itemsep=0pt, parsep=0pt, partopsep=0pt}
\setlist[enumerate]{leftmargin=*, topsep=.0em, itemsep=0pt, parsep=0pt, partopsep=0pt}

\usepackage{amsthm}
\newtheoremstyle{def_style}
  {0.5em}      
  {0.5em}      
  {}          
  {}          
  {\bfseries} 
  {.}         
  {0.5em}      
  {}          

\theoremstyle{def_style}

\theoremstyle{def_style}

\usepackage{amsmath}
\usepackage{bbm}

\usepackage{collectbox}



\usepackage{mathtools}

\usepackage{balance}

\usepackage{caption}
\usepackage{subcaption}

%% file: section/author.tex
\author{Jiarui Sun}
\authornote{Work done while at Visa Research.}
\affiliation{%
  \institution{University of Illinois Urbana-Champaign}
  \city{Urbana}
  \state{IL}
  \country{USA}
}

\author{Chin-Chia Michael Yeh}
\author{Yujie Fan}
\author{Xin Dai}
\affiliation{%
  \institution{Visa Research}
  \city{Foster City}
  \state{CA}
  \country{USA}
}

\author{Xiran Fan}
\author{Zhimeng Jiang}
\author{Uday Singh Saini}
\affiliation{%
  \institution{Visa Research}
  \city{Foster City}
  \state{CA}
  \country{USA}
}

\author{Vivian Lai}
\author{Junpeng Wang}
\author{Huiyuan Chen}
\affiliation{%
  \institution{Visa Research}
  \city{Foster City}
  \state{CA}
  \country{USA}
}

\author{Zhongfang Zhuang}
\author{Yan Zheng}
\affiliation{%
  \institution{Visa Research}
  \city{Foster City}
  \state{CA}
  \country{USA}
}

\author{Girish Chowdhary}
\affiliation{%
  \institution{University of Illinois Urbana-Champaign}
  \city{Urbana}
  \state{IL}
  \country{USA}
}

%% file: section/abstract.tex
\begin{abstract}

Time series forecasting at scale presents significant challenges for modern prediction systems, particularly when dealing with large sets of synchronized series, such as in a global payment network.
In such systems, three key challenges must be overcome for accurate and scalable predictions: 1) emergence of new entities, 2) disappearance of existing entities, and 3) the large number of entities present in the data.
The recently proposed Inverted Transformer (iTransformer) architecture has shown promising results by effectively handling variable entities.
However, its practical application in large-scale settings is limited by quadratic time and space complexity ($O(N^2)$) with respect to the number of entities $N$.
In this paper, we introduce \name{}, an improved inverted transformer architecture that maintains the adaptive capabilities of iTransformer while reducing computational complexity to linear scale ($O(N)$).
Our key innovation lies in restructuring the attention mechanism to eliminate redundant computations without sacrificing model expressiveness.
Additionally, we incorporate a random projection mechanism that not only enhances efficiency but also improves prediction accuracy through better feature representation.
Extensive experiments on the public LargeST benchmark dataset and a proprietary large-scale time series dataset demonstrate that \name{} significantly outperforms existing methods in both computational efficiency and forecasting accuracy.
Our approach enables practical deployment of transformer-based forecasting in industrial applications where handling time series at scale is essential.

\vspace{-0.4mm}
\end{abstract}

%% file: section/introduction.tex
\section{Introduction}

\input{insert/fig_emerge_vanish}

For companies operating a global payment network, spatial-temporal time series data is a significant type of information. 
Accurately forecasting \textit{future} transaction volumes could provide considerable advantages to entities involved in the payment process, serving as a strategic tool for decision-making processes, risk management, and resource allocation. 
Within this complex network, entities are interconnected, each one reciprocally influencing the others. 
The collective transaction volumes over time constitute more than a simple time series dataset. 
Instead, they create a \textit{large-scale spatial-temporal time series dataset}, a rich and intricate network resulting from the interactions of millions of entities. 
This dataset encapsulates not only the temporal dynamics of transaction volumes but also the spatial relationships among these entities. 
Extracting the predictive power from this dataset could facilitate more effective and efficient operations within the payment network.

This large-scale spatial-temporal time series dataset from payment networks presents two major challenges. 
Firstly, the spatial-temporal forecasting model must be scalable to accommodate the millions of entities present in the payment network. 
Secondly, the model needs to deal with entities that emerge and vanish as they participate in payment activities. 
As shown in \cref{fig:emerge_vanish}, our dataset contains many instances of both emerged and vanished entities.

Several solutions have been proposed to tackle the first challenge. 
Notably, mixer-based models such as RPMixer~\cite{yeh2024rpmixer} employ a mixer layer, or more precisely a \textit{feature multi-layer perceptron (MLP)}, as illustrated in \cref{fig:overview} (5), to model the relationships among entities. 
This stands in contrast to many spatial-temporal forecasting models which explicitly compute the correlation matrix~\cite{shang2021discrete,lan2022dstagnn,shao2022pre,liu2023itransformer}, as depicted in \cref{fig:overview} (4). 
Consequently, mixer-based models exhibit linear time/space complexity \wrt the number of entities, a stark contrast to the quadratic complexity found in other models~\cite{shang2021discrete,lan2022dstagnn,shao2022pre,liu2023itransformer}. 
Despite their efficiency, these mixer-based models fall short in addressing the second challenge. 
As seen in \cref{fig:overview} (5), they utilize a fixed feature MLP to capture the correlation among entities. 
When these correlations change due to the emergence or disappearance of entities, the feature MLP might generate inaccurate predictions. 
Specifically, predictions for an emerging entity could be incorrect as the feature MLP does not consider the correlation between the new and existing entities. 
Likewise, when an entity disappears, the predictability for entities that were strongly correlated with the disappeared entity may be adversely affected due to the unusual activity (\ie, consecutive zeros) associated with the vanished entity.

\input{insert/fig_overview}

Conversely, methods such as iTransformer~\cite{liu2023itransformer} tackle the second challenge by taking advantage of the inductive bias inherent in the attention mechanism. 
Specifically, the attention mechanism identifies the relationships among various entities based on the most recent time series. 
This implies that the correlations perceived by the model can adapt to both emerging and vanishing entities. 
For emerging entities, the resulting correlation with other entities will mirror those of existing entities exhibiting similar temporal patterns. 
For vanishing entities, their unusual temporal patterns (\ie, consecutive zeros) are disregarded when making predictions for other entities. 
However, despite these advantages, the method suffers from poor scalability due to the necessity of computing the entity correlation matrix, \ie, \cref{fig:overview} (4), \wrt the number of nodes (\ie, quadratic), making it unsuitable for use in payment networks.

In this paper, we introduce an efficient method termed Efficient iTransformer (\ie, \name{}), which has linear time/space complexity \wrt the number of entities, akin to RPMixer~\cite{yeh2024rpmixer}, whilst preserving the inductive bias innate to the attention mechanism. 
Additionally, we incorporate the random projection design~\cite{yeh2024rpmixer} into \name{} to further enhance its performance in handling large-scale spatial-temporal forecasting problems.

Our key contributions are:

\begin{itemize} 
    \item Proposal of a scalable method, \name{}, capable of effectively managing emergent and vanishing entities. 
    \item Design of three benchmark experiment settings to evaluate a method's ability to handle large-scale spatial-temporal time series data with emerging and vanishing entities, using the publicly available LargeST dataset~\cite{liu2024largest}. 
    \item Demonstration of the scalability and accuracy of the proposed \name{} in comparison with alternative methods, validated using both emulated and real datasets. 
\end{itemize}


%% file: insert/fig_emerge_vanish.tex
\begin{figure}[htp]
\centerline{
\includegraphics[width=0.8\linewidth]{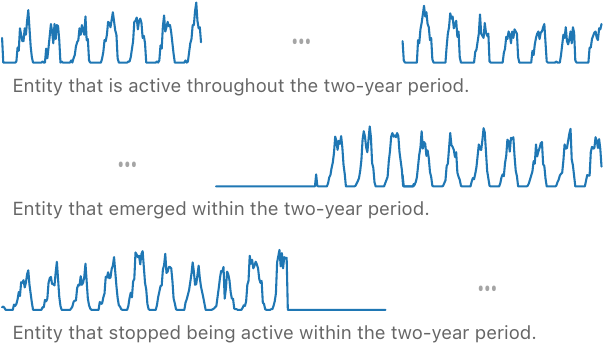}
}
\caption{
Of the 23,849 entities sampled from our network over a two-year period, 14,252 remained active throughout the duration. 
Meanwhile, 1,974 entities emerged and were not present during the first week of the sampled period. 
In addition, 5,880 entities ceased activity before the final week of the period.}
\label{fig:emerge_vanish}
\end{figure}

%% file: insert/fig_overview.tex
\begin{figure*}[t]
\centerline{
\includegraphics[width=0.75\linewidth]{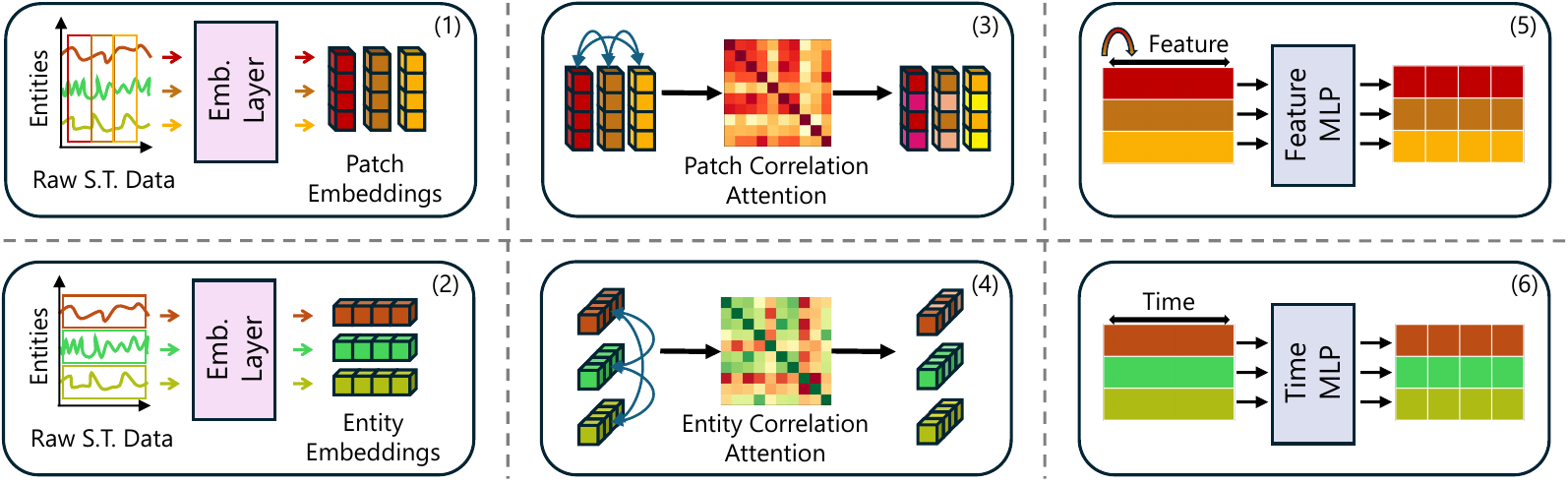}
}
\caption{
Different spatial-temporal modeling designs. (1, 2): patch vs. entity embedding. (3, 4): temporal patch vs. spatial entity attention. (5, 6): feature vs. time MLP.
PatchTST \cite{nie2022time}: (1) + (3) + (5); iTransformer \cite{liu2023itransformer}: (2) + (4) + (6); TSMixer \cite{chen2023tsmixer}: (6) + (5).
}
\label{fig:overview}
\end{figure*}

%% file: section/related.tex
\section{Related Work}

\subsection{Spatial-Temporal Forecasting}

Spatial-temporal forecasting is a critical research area driven by ubiquitous real-world data, such as motion \cite{sun2024comusion, sun2023towards} and traffic \cite{sun2023revealing, bui2022spatial}, that exhibits spatial and temporal dependencies. 
Effectively capturing these complex dependencies requires learning informative representations, a challenge that has been central to success across various domains \cite{jiao2023learning, jiao2022fine, chen2023improving, sun2024mooss, liu2025learning}. 
To address this representation challenge in spatial-temporal forecasting, researchers have developed methods that typically model spatial relations in one of two ways: 1) grids \cite{zhang2023mlpst}, where feature maps are split spatially into non-overlapping equal-sized patches, or 2) graphs \cite{liu2024largest}, where entities are represented as nodes with connections between them. 
While both approaches have their merits, graph-based methods have gained significant attention due to their flexibility in modeling irregular structures.
These methods rely on graph neural networks (GNN) \cite{li2018dcrnn_traffic, wu2020connecting, bai2020adaptive, sun2023revealing, wu2019graphwave, guo2019attention, yu2018spatio, fang2021spatial, li2023dgcrn, lan2022dstagnn, shang2024transitivity, shao2022d2stgnn}, Transformers \cite{xu2020spatial} or multilayer perceptrons (MLP) \cite{yeh2024rpmixer, ijcai2022p111} to capture spatial dependencies among nodes, and temporally, they leverage recurrent neural networks (RNN), temporal convolutional networks (TCN) or Transformers for sequence modeling.
Despite their effectiveness, these approaches still face challenges in scaling to very large spatial-temporal data such as a payment network, and adapting to complex real-world cases where nodes/entities may appear or disappear over time, issues that our work aims to address.

\subsection{Multivariate Time Series Forecasting}
The task of spatial-temporal forecasting is often considered as multivariate time series (MTS) forecasting when the spatial input graph is absent or not clearly defined.
In this line of work, Transformer-based methods \cite{li2019enhancing, liu2021pyraformer, wu2021autoformer, zhou2021informer, zhou2022fedformer} have emerged due to their sequence modeling ability.
Recently, approaches such as \cite{zeng2023transformers, liu2023itransformer, chen2023tsmixer, yeh2024rpmixer} have become popular due to their linear complexity \wrt sequence length and forecasting performance.
Our work builds upon these efficient approaches, proposing \name{}, an improved iTransformer architecture that significantly enhances efficiency while maintaining forecasting capability.

\subsection{Scalable Forecasting Methods}
Scalability in forecasting methods is crucial for modeling large-scale datasets \cite{liu2024largest} containing numerous, varying number of entities per sample and for performing long-term, multi-step predictions.
While Transformer-based models offer strong performance, they typically suffer from quadratic complexity \wrt sequence length. 
Recent approaches like mixer-based and linear models achieve linear time complexity for sequence length, but they still failed to handle a dynamic number of entities which might occur during both training and testing phases. 
The iTransformer model \cite{liu2023itransformer} effectively handles new entities but still incurs quadratic complexity in the number of nodes. 
Our proposed method addresses these limitations by achieving linear complexity in both space and time while remaining inductive to the number of entities, thus providing a more comprehensive scalable solution for forecasting problems.

%% file: section/background.tex
\section{Definition and Problem}

\begin{figure}[t]
\centerline{
\includegraphics[width=0.8\linewidth]{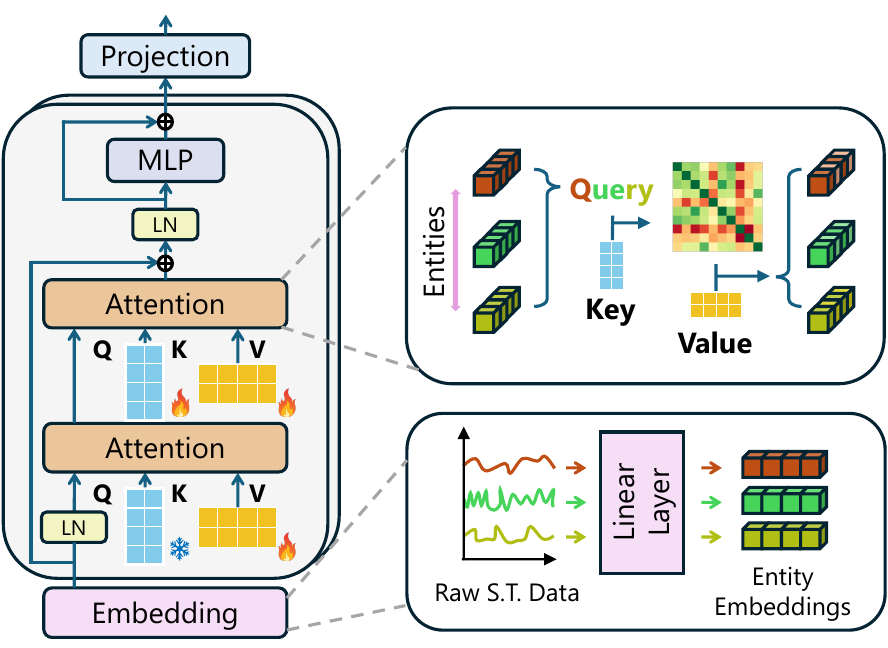}
}
\caption{
The proposed \name{} architecture. The randomly initialized key matrix of the initial attention layer of each Efficient iTransformer Module is frozen, serving as the random projection operator.
}
\label{fig:model}
\end{figure}

In spatial-temporal forecasting, given historical observation $\rmX \in \sR^{H\times N\times C}$ containing $H$ frames of $N$ time-dependent variables with $C$ features, our objective is to predict the the subsequent value $\rmY \in \sR^{F\times N\times C}$ of the $F$ following steps.
All $N$ variables in $\rmX$ are not only temporally evolving but also structurally interrelated. 
This structural dependency can be represented by a graph $\gG = (\gV, \gE, \rmA_\text{adj})$, where $\gV$ and $\gE$ are the sets of $N$ variables and their relations respectively, $\rmA_{\text{adj}} \in \sR^{N\times N}$ denotes adjacency matrix representing variables' connectivity.
Consequently, the spatial-temporal forecasting problem can be formally defined as:
\begin{equation}
    f_{\theta}(\rmX, \rmA_{\text{adj}}) \rightarrow \rmY,
\end{equation}
where $f_{\theta}(\cdot)$ denotes the parameterized forecaster.
We note that some works, such as \cite{li2018dcrnn_traffic}, rely on a predefined $\gG$ as part of the model input, while others, like \cite{wu2020connecting}, aim to learn the structural dependencies between variables by constructing $\gG$ during the learning process. 
The iTransformer model \cite{liu2023itransformer} belongs to the latter category. 
By leveraging the self-attention mechanism on the spatial dimension, it can handle varying numbers of entities across different learning stages. 
However, this flexibility comes at the cost of quadratic complexity in the number of entities.

%% file: section/method.tex
\section{Proposed Method}

\begin{figure}[t]
\centerline{
\includegraphics[width=0.85\linewidth]{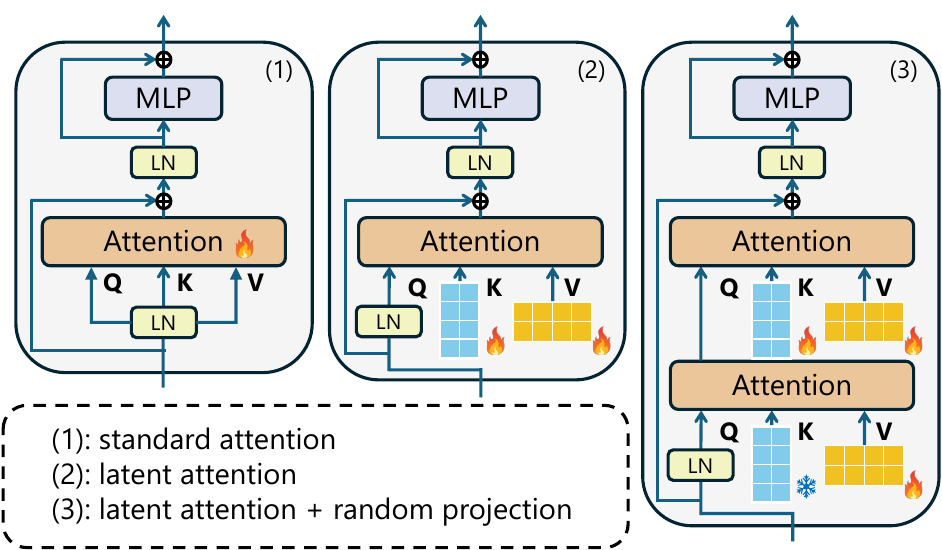}
}
\caption{
Design progression of \name.
}
\label{fig:improve}
\end{figure}

Our proposed~\name{}, as shown in \cref{fig:model}, is designed to improve the forecasting performance of iTransformer in terms of efficiency for large-scale spatial-temporal data, while maintaining the predictive capability.
It consists of two novel components: Efficient iTransformer Module (EiM) and Random Projection Module (RP).

\subsection{Efficient iTransformer Module}

\paragraph{Entity-level embedding.} 
With the extensive success of Transformers on domains of language and vision, these models have been adapted to forecasting task to capture data dependencies. 
Before iTransformer, the primary focus of the sequence modeling power of Transformers in MTS forecasting was on the \textit{temporal} dimension, treating multiple variates of the same time step as parts of one single token, as shown in \cref{fig:overview} (1). 
In contrast, our approach adopts iTransformer's orthogonal embedding method  (\cref{fig:overview} (2)), where each entity (\eg, sensor, payment node) is represented as a separate token across all time steps. 
This allows models to generalize to varying numbers of entities they process.
Recall that $\rmX$ is the observation with $T$ time steps, $N$ entities, and $C$ channels. 
Formally, we have the embedding function $\text{Embedding}( \sR^{T \times C} \mapsto \sR^{D})$:
\begin{equation}
    \rvh_n^0 =\text{Embedding}\left(\rmX_{:,n,:}\right), \rvh_n^0 \in \sR^{D},
\end{equation}
where $D$ is the embedding dimension, and $\text{Embedding}(\cdot)$ is implemented as a simple linear layer.
The packed embedding for all historical observation is then $\rmH^0 \in \sR^{N \times D}$.

\paragraph{Efficient latent attention.}
By embedding each time series as a whole, we follow iTransformer's approach of using variant-wise attention to learn entity correlations. 
While this method in \cref{fig:overview} (3) is inductive to the number of entities, it incurs an quadratic complexity $O(N^2)$. 
To improve efficiency, we propose an efficient latent attention mechanism with fixed-size latent factors.
Let $\rmH  \in \sR^{N \times D}$ be the input to latent attention after being layer normalized. Formally, we have:
\begin{align}
    & \rmQ = \rmH \rmW, \rmK \in \sR^{M \times D}, \rmV \in \sR^{M \times D}, \\
    & \text{Latent\_Att}(\rmH) = \text{softmax}(\rmQ\rmK^\top/\sqrt{D})\rmV,
\label{eq:lta}
\end{align}
where $\rmW \in \sR^{D\times D}, \rmK \in \sR^{M \times D}, \rmV \in \sR^{M \times D}$ are learnable matrices.
Notably, the attention map $\rmA_{\text{att}} \in \sR^{N \times M} \propto \text{softmax}(\rmQ\rmK^\top)$ produced by the latent attention has an input-independent dimension $M$.
Our key insight is that structural characteristics are often shared across different time series, suggesting that a set of latent factors with fixed size can effectively capture the essential inter-entity relationships.
By leveraging latent factors with input-independent size, our mechanism reduces computational complexity from quadratic to linear, while maintaining the ability to model variant dependencies, striking a balance between efficiency and expressiveness.

\paragraph{Temporal MLP}
With latent attention handling inter-entity modeling, the task of temporal modeling for \name{} is delegated to temporal MLP blocks.
This approach aligns with recent findings \cite{zeng2023transformers, chen2023tsmixer, das2023long} demonstrating the effectiveness of simpler, linear models in temporal modeling.
Let $\rmH \in \sR^{N \times D}$ denote the layer-normalized input to the temporal MLP.
We define:
\begin{equation}
    \operatorname{MLP_{\text{temp}}}\left(\rmH\right) = \text{MLP}(\rmH),
\end{equation}
where we leverage feed-forward layers to learn the series representation of each entity across the temporal dimension.

\subsection{Random Projection Mechanism}
While the above Efficient iTransformer Module shown in \cref{fig:improve} (2) effectively and inductively captures spatial-temporal relationships with linear complexity, the sheer scale of real-world datasets, such as the payment network we are interested in modeling, present additional challenges. 
Specifically, large-scale datasets often exhibit intricate node relationships that are difficult to capture fully, and the risk of overfitting to dominant patterns is significant.
To address these, inspired by \cite{yeh2024rpmixer}, we adopt a Random Projection Mechanism that enhances \name{}'s capacity to learn diverse representations, as shown in \cref{fig:improve} (3).
Specifically, following \cite{yeh2024rpmixer}'s two-level spatial-modeling structure, after we obtain the entity-level embeddings $\rmH^0$, we first pass it to an efficient latent attention module with the $\rmK$ matrix frozen, whose output is then modeled with fully learnable latent attention and temporal MLP. That is:
\begin{equation}
    \rmQ = \rmH^0\rmW, \text{Random\_Proj}(\rmH^0) = \text{softmax}(\rmQ\rmK_\text{frozen}^\top/\sqrt{D})\rmV,
\label{eq:rpm}
\end{equation}
Specifically, having a frozen random $\rmK$ produces a randomized $\rmQ\rmK^\top$ in \cref{eq:lta}, which describes the attention between each entity and the latent factors.
This randomized attention mechanism functions similarly to the random projection layer in \cite{yeh2024rpmixer}, as it effectively combines the rows of $\rmV$ in a predetermined manner, with each row corresponding to a distinct latent factor.
Such mechanism first acts as a noise-equivalent regularization to mitigate model overfitting.
It also encourages different \name{} blocks to concentrate on distinct sets of entities, thereby producing diverse representations which are then further modeled by the subsequent fully-learnable latent attention layer.
This diversity boosts \name{}'s ability to capture a wide range of spatial-temporal patterns. 

%% file: section/experiment.tex
\section{Experiment}
The source code for the experiments associated with the publicly available datasets can be downloaded from~\cite{supplementary}.

\subsection{Datasets, Baselines and Metrics}

\paragraph{Datasets.}
To assess the effectiveness of \name, we first utilize the recently published LargeST dataset \cite{liu2024largest}. 
From this comprehensive dataset, we generated four sub-datasets: SD (San Diego), GBA (Greater Bay Area), GLA (Greater Los Angeles), and CA (California). 
The statistics for these sub-datasets are presented in \cref{fig:dataset}.
Following the experimental setup described in \cite{liu2024largest}, we chronologically divided each sub-dataset into training, validation, and test sets using a 6:2:2 ratio. 
The forecasting task involves forecasting the next 12 time steps based on historical traffic sensory data.
In addition to the standard case (Scenario 0) where node composition remains unchanged, we create three simulated scenarios from the LargeST dataset~\cite{liu2024largest} to emulate real-world dynamic conditions:
\begin{enumerate}
\item Scenario 1: 10\% of nodes in the test set are new, simulating the emergence of new entities.
\item Scenario 2: 10\% of nodes from the training set are absent in the test set, simulating the disappearance of entities.
\item Scenario 3: A combination of Scenarios 1 and 2, with 10\% new entities in the test set and 10\% of training set entities missing from the test set.
\end{enumerate}

While the LargeST dataset serves as a proxy for real transaction data, we have also evaluated the proposed method using spatial-temporal time series data derived from an actual payment network. 
Our transaction spatial-temporal time series dataset comprises per-hour transaction volume time series for two years, sampled from 23,849 interconnected entities within our network. 
The dataset was chronologically divided into training, validation, and test sets following a 6:2:2 ratio. 
The forecasting task involves predicting the transaction volume for the forthcoming 7 days (or 168 time steps/hours) based on historical transaction volume data.

\paragraph{Baselines.}

We evaluate \name{} against TSMixer \cite{chen2023tsmixer}, RPMixer \cite{yeh2024rpmixer}, and iTransformer \cite{liu2023itransformer}, along with their variants incorporating our proposed efficient attention (EiM) and random projection (RP) modules.
It is important to note that we focus on comparing methods that do not utilize graph structures, as our dataset does not include graph information.
We also include a linear model baseline as a performance benchmark and sanity check.
\paragraph{Metrics. }
In terms of performance evaluation, we adopt three metrics, including mean absolute error (MAE), root mean squared error (RMSE), and mean absolute percentage error (MAPE).
These metrics are commonly used in evaluating spatial-temporal/MTS forecasting tasks.

\subsection{LargeST: Semi-Synthetic Dataset Result}

\input{insert/tab_largest_simulation}

\cref{tab:largest_simulation} presents our experimental results on the LargeST dataset, serving as both a performance comparison and an ablation study for the proposed EiM and RP modules.
Through analyzing the results, we have the following observations:
1) On the smallest dataset (SD), iTransformer with Random Projection (iXFMR+RP) excels, particularly in Scenarios 1, 2, and 3 where node composition changes, highlighting the importance of inductiveness \wrt number of entities and the robustness brought by random projection.
However, the applicability of iTransformer is limited to this small-sized dataset due to scalability issues.
2) In the standard case (Scenario 0) with no node changes, the proposed EiM enables iTransformer to process larger datasets effectively, achieving the best performance across most metrics.
3) Most importantly, as we progress to larger datasets and more complex scenarios (CA is over ten times larger than SD as shown in \cref{fig:dataset}), \name{} with both EiM and RP enabled demonstrates increasing advantages. 
The performance gap widens in favor of \name{} for the most challenging cases, showcasing its effectiveness in handling large-scale, dynamic scenarios.
These findings underscore the efficiency and effectiveness of our approach.

\input{insert/fig_dataset}

\subsection{Transaction Dataset Result}
\cref{tab:transaction} presents the quantitative results on our real-world proprietary transaction volume dataset.
\input{insert/tab_transaction}
From the results, we can first confirm that \name{} consistently outperforms baseline models across all forecasting horizons, even on our challenging, large-scale transaction prediction task. 
We first note that the seemingly high errors (\eg, RMSE metric) stem from the inherent challenges of our dataset, that  1) our focus on high-traffic merchants, where hourly transactions can reach millions, results in large absolute errors; 2) the extended prediction horizon of 168 steps ahead (compared to typical traffic forecasting horizons of 12 steps) amplifies error accumulation.
For the 1-day, 3-day, and 7-day horizons, \name{} shows the best performance across all metrics. 
This consistent performance underscores \name{}'s robustness in modeling complex spatial-temporal behaviors from real-world scenarios.
Second, we can observe that compared with results obtained from LargeST dataset, \name{} exhibits significantly larger improvements over all baselines on the transaction dataset. 
On average across all horizons, \name{} improves upon the best performing baseline by \textbf{24.6\%} in MAE, \textbf{33.0\%} in RMSE, and \textbf{11.2\%} in MAPE.
These substantial gains further highlight the effectiveness of our approach in handling complex, large-scale transaction data, validating its potential for practical applications.

\input{insert/fig_qualitative}

To better comprehend how \name{} surpasses alternative solutions, we visualized the predicted time series for \name{} and the more competitive alternatives, TSMixer+EiM, RPMixer+EiM, and iXFMR+EiM. 
The qualitative visualization can be seen in \cref{fig:qualitative}, which highlights several challenges associated with real-world large-scale transaction data. 
For entities 0, 11117, and 13635, the transaction volume increases during the tested time steps compared to the historical time steps. 
Conversely, entities 13644 and 12480 exhibit decreasing or vanishing transaction volumes. 
For entity 13650, a change in the weekly pattern is observable. 
It is evident that our proposed method \name{} can adapt to these changes more swiftly than the alternative methods, which elucidates why \name{} outperforms the alternatives quantitatively.

\subsection{Network Architecture Analysis}
In this section we analyze the differences between \name{} and iXFMR+EiM via the lens of representation similarity measure Linear CKA\cite{DBLP:journals/corr/abs-1905-00414} to demonstrate the effectiveness of random projection. Next, we briefly lay out the experimental protocol but we refer the readers to \cite{klabunde2024similarityneuralnetworkmodels} for a survey of other representational similarity measures for a more detailed study. Borrowing the notation from \cite{klabunde2024similarityneuralnetworkmodels}, given an input $\mathbf{X} \in \mathbb{R}^{M \times N \times L}$, where $N$ and $L$ are the number of spatiotemporal features and time horizon of $M$ independent time series respectively and a model $f_{\theta}(\cdot)$. We obtain the internal representations of the data as follows:
\begin{equation}
    \mathbf{D}^{(l)} = \left(f^{(l)}_{\theta} \circ f^{(l-1)}_{\theta} \circ \dots \circ f^{(1)}_{\theta}\right)(\mathbf{X})\in\mathbb{R}^{M \times N \times F} \forall l.
    \label{eq:rep_extract}
\end{equation}

All extracted representations $\mathbf{D}^{(l)}\in\mathbb{R}^{M \times N \times F} $ for each network layer $l$ are then suitably flattened so that $\mathbf{D}^{(l)}\in\mathbb{R}^{M \times NF}$. Then for 2 layers $p,q$ of a network (or different networks) we compute their similarity as $\textit{CKA}(\mathbf{C}^{(p)},\mathbf{C}^{(q)})$ based on \autoref{eq:CKA}. Where  $\mathbf{C}^{(p)}=\mathbf{D}^{(p)}\mathbf{D}^{(p)T} \in\mathbb{R}^{M \times M} $ and $\textit{HSIC}(\mathbf{C}^{(p)},\mathbf{C}^{(q)})$ stands for the Hilbert-Schmidt Independence Criterion \cite{NIPS2007_d5cfead9} and $H = I_{M} - \frac{1}{M}\textbf{11}^{T} \in\mathbb{R}^{M \times M}$ is a centering matrix. 

\begin{equation}
    \textit{CKA}(\mathbf{C}^{(p)},\mathbf{C}^{(q)}) = \frac{\textit{HSIC}(\mathbf{C}^{(p)},\mathbf{C}^{(q)})}{\sqrt{\textit{HSIC}(\mathbf{C}^{(p)},\mathbf{C}^{(p)}) \textit{HSIC}(\mathbf{C}^{(q)},\mathbf{C}^{(q)})}}
    \label{eq:CKA}
\end{equation}

\begin{equation}
    \textit{HSIC}(\mathbf{C}^{(p)},\mathbf{C}^{(q)}) = \frac{\textit{trace}(H\mathbf{C}^{(p)}HH\mathbf{C}^{(q)}H)}{(M-1)^{2}}
    \label{eq:HSIC}
\end{equation}

\input{insert/fig_net_analysis_linear_cka}

Based on the earlier description, we compute pairwise layer similarity scores for \name{} and iXFMR+EiM as well compare them head to head, the results of which are shown in \autoref{fig:Linear_CKA_EiF_IXFMR-EiM}. In \autoref{fig:CKA_EiF} we show the computed scores for \name{}, \autoref{fig:CKA_IXFMR-EiM} shows the same scores for iXFMR+EiM and \autoref{fig:CKA_EiF_IXFMR-EiM} shows the scores when comparing layers of \name{} directly with iXFMR+EiM. We observe that unlike iXFMR+EiM which has a higher degree of similarity in its learned representations through the layers, \name{} on the other shows a degree of evolution in its internal representations which lead to the later stages of the model learning distinct representation from its earlier layers thereby demonstrating differences in the representation learning mechanisms of the 2 architectures.

\subsection{Runtime and Memory Cost Analysis}


In this section, we empirically demonstrate the efficiency advantages of \name{} over baselines and their variants. 
To evaluate computational cost, we vary the number of entities in the spatial-temporal data instance from $10^0$ to $10^6$, performing standard 12-step ahead prediction using a single data sample.

First, As shown in the upper plot of \cref{fig:cost_forward}, \name{} exhibits near-linear empirical runtime scaling, with performance slightly below the Linear baseline but superior to other methods. 
Mixer-based methods show runtime efficiency degradation beyond $10^5$ entities, while the original iTransformer encounters out-of-memory (OOM) issue at approximately 10,000 entities. 

Second, regarding memory consumption, illustrated in the lower plot of \cref{fig:cost_forward}, \name{} successfully mitigates the memory bottleneck of the original iTransformer, preventing OOM issue for large entity counts. 
Although \name{}'s empirical memory profile is higher than Mixer-based models, this is attributable to the larger number of forward activations that must be retained during the forward pass due to the extensive attention-based interactions, which enables its superior predictive performance.
These results demonstrate that \name{} achieves a favorable balance between computational efficiency and model expressiveness, making it well-suited for dynamic, large-scale spatial-temporal forecasting task.

\input{insert/fig_cost_forward}

\subsection{Parameter Sensitivity Analysis}

We conduct a hyper-parameter sensitivity analysis on the two most promising methods from \cref{tab:largest_simulation}: iTransformer with efficient latent attention (iXFMR+EiM) and \name{}, using the SD dataset from LargeST. 
As shown in \cref{fig:param_lr,fig:param_layer,fig:param_neuron}, both models benefit from larger model sizes with an optimal learning rate of $10^{-4}$. 
Notably, \name{} not only shows better MAE performance, but also demonstrates lower sensitivity to hyper-parameter settings compared to iXFMR+EiM. 
This reduced sensitivity underscores the importance of \name{}'s RP Module, which generates diverse representations and thereby improves adaptability across different task settings. 
As such, \name{} is particularly suitable for large-scale spatial-temporal forecasting, as it maintains consistent performance across various settings thus reduces tuning efforts.

\input{insert/fig_param_lr}

\input{insert/fig_param_layer}

\input{insert/fig_param_neuron}

\subsection{LargeST: Benchmark Result}
For completeness, we present benchmark results in \cref{tab:largest_benchmark} on LargeST, comparing \name{} with established spatial-temporal baselines. 
Our experiments reveal that RPMixer exhibits strong performance in simpler cases, while the iXFMR+EiM variant shows advantages in more challenging scenarios. 
Similar to the synthesis dataset result, \name{} demonstrates superior performance on the CA dataset, which highlights our model's scalability and effectiveness in modeling dynamic, intricate spatial-temporal dependencies. 

\input{insert/tab_largest_benchmark}

%% file: insert/tab_largest_simulation.tex
\begin{table*}[pt]
\caption{Performance comparisons on LargeST.
We bold the best-performing results.
The performance reported is computed by averaging over 12 predicted time steps.
The absence of iTransformer (iXFMR) and its RP variant on GBA, GLA and CA indicates that it incurs out-of-memory issue.}

\label{tab:largest_simulation}
\begin{center}
\resizebox{0.75\linewidth}{!}{%
\begin{tabular}{llccc|ccc|ccc|ccc}
    \shline
    \multirow{2}{*}{Data} & \multirow{2}{*}{Method}  & \multicolumn{3}{c}{Scenario 0} & \multicolumn{3}{c}{Scenario 1} & \multicolumn{3}{c}{Scenario 2} & \multicolumn{3}{c}{Scenario 3} \\ \cline{3-14} 
     &  & MAE & RMSE & MAPE & MAE & RMSE & MAPE & MAE & RMSE & MAPE & MAE & RMSE & MAPE \\ 
     \hline \hline
     \multirow{8}{*}{SD} & Linear & 25.85 & 42.35 & 17.10\% & 25.81 & 42.33 & 16.84\% & 25.63 & 41.89 & 16.65\% & 25.83 & 42.35  & 17.14\% \\
     & TSMixer    & 19.06 & 30.66 & 12.55\% & 32.83 & 79.56 & 30.65\% & 18.57 & 32.71 & 12.76\% & 29.47 & 111.48 & 34.28\% \\
     & RPMixer  & \textbf{16.90} & \textbf{27.97} & 11.07\% & 23.05 & 40.07 & 16.43\% & 18.96 & 30.91 & 12.43\% & 22.67 & 39.20  & 17.77\% \\
     & iXFMR    & 17.34 & 29.06 & 10.93\% & 17.66 & 29.50 & 11.12\% & \textbf{17.23} & \textbf{29.09} & 10.69\% & 17.42 & 29.31  & 11.05\% \\
     \cline{2-14}
     & TSMixer+EiM  & 18.89 & 31.89 & 12.24\% & 19.25 & 32.40 & 12.65\% & 19.02 & 31.93 & 12.23\% & 20.83 & 34.36  & 14.25\% \\
     & RPMixer+EiM  & 18.13 & 30.63 & 11.54\%  & 18.30 & 31.05 & 11.64\%  & 18.10 & 30.73 & 11.21\% & 18.20 & 30.91 & 11.47\%  \\
     & iXFMR+EiM   & 17.72 & 30.24 & 11.03\% & 17.73 & 30.52 & 11.10\% & 17.85 & 30.52 & 10.98\% & 17.87 & 30.62  & 11.23\% \\
     & iXFMR+RP   & 17.22 & 28.86 & \textbf{10.91}\% & \textbf{17.35} & \textbf{29.10} & \textbf{10.97}\% & 17.27 & 29.16 & \textbf{10.67}\% & \textbf{17.28} & \textbf{29.03} & \textbf{10.87}\% \\
     \cline{2-14}
     & \name{} (Ours)   & 17.70 & 30.10 & 11.09\% & 17.91 & 30.61 & 11.25\% & 17.71 & 30.20 & 10.99\% & 17.79 & 30.61  & 11.24\% \\
     \hline \hline
     \multirow{7}{*}{GBA} & Linear  & 26.12 & 42.14 & 22.10\% & 26.07 & 42.26 & 21.56\% & 26.11 & 42.33 & 21.56\% & 26.06 & 42.28 & 21.58\% \\
     & TSMixer  & 19.58 & 32.56 & 16.58\% & 27.65 & 84.84 & 25.24\% & 20.35 & 33.02 & 17.03\% & 26.49 & 69.85 & 29.07\% \\
     & RPMixer   & 19.06 & \textbf{31.54} & 15.09\% & 22.94 & 36.93 & 20.95\% & 20.26 & 33.89 & 17.19\% & 23.06 & 37.55 & 20.88\% \\
     \cline{2-14}
     & TSMixer+EiM  & 19.68 & 33.20 & 15.65\% & 19.60 & 32.97 & 15.48\% & 22.09 & 35.79 & 19.98\% & 19.71 & 33.23 & 15.87\% \\
     & RPMixer+EiM  & 18.98 & 32.62 & 14.58\% & 19.16 & 32.93 & 14.93\% & 19.20 & 32.85 & 14.93\% & 19.08 & 32.70 & 14.58\% \\
     & iXFMR+EiM   & \textbf{18.72} & 32.58 & \textbf{13.55}\% & 18.83 & \textbf{32.65} & 14.15\% & \textbf{18.85} & \textbf{32.71} & 14.06\% & 18.99 & 32.76 & 14.41\% \\
     \cline{2-14}
     & \name{} (Ours)  & 18.76 & 32.59 & 13.65\% & \textbf{18.82} & 32.81 & \textbf{14.13}\% & 18.92 & 32.79 & \textbf{13.95}\% & \textbf{18.72} & \textbf{32.59} & \textbf{13.79}\% \\
     \hline \hline
     \multirow{7}{*}{GLA} & Linear  & 26.40 & 42.56  & 17.16\% & 26.47 & 42.62 & 17.19\% & 26.45 & 42.64 & 17.22\% & 26.40 & 42.56  & 16.97\% \\
     & TSMixer  & 22.12 & 207.68 & 14.87\% & 31.50 & 95.35 & 27.55\% & 20.91 & 90.29 & 13.89\% & 31.44 & 133.27 & 32.26\% \\
     & RPMixer  & 18.46 & \textbf{30.13}  & 11.34\% & 24.02 & 39.31 & 17.51\% & 19.95 & 31.88 & 13.45\% & 23.81 & 39.17  & 16.92\% \\
     \cline{2-14}
     & TSMixer+EiM & 19.76 & 32.95  & 12.46\% & 19.38 & 32.41 & 11.61\% & 19.40 & 32.36 & 11.66\% & 19.38 & 32.42  & 11.99\% \\
     & RPMixer+EiM & 18.60 & 31.40  & 11.22\% & 18.65 & 31.53 & 11.19\% & 18.71 & 31.56 & 11.23\% & 18.88 & 31.82  & 11.37\% \\
     & iXFMR+EiM  & \textbf{18.14} & 30.88  & \textbf{10.76}\% & \textbf{18.22} &\textbf{ 31.04} & 10.85\% & 18.32 & 31.12 & 10.93\% & 18.39 & 31.13  & 10.99\% \\
     \cline{2-14}
     & \name{} (Ours)  & 18.22 & 31.17  & 10.80\% & 18.23 & 31.07 & \textbf{10.77}\% &\textbf{ 18.22} &\textbf{ 31.11} & \textbf{10.65}\% & \textbf{18.17} & \textbf{31.03 } & \textbf{10.70}\% \\
     \hline \hline
     \multirow{7}{*}{CA} & Linear  & 24.32 & 39.88 & 18.52\% & 24.32 & 39.90 & 18.59\% & 24.35 & 39.91 & 18.74\% & 24.36 & 39.91  & 18.55\% \\
     & TSMixer& 19.86 & 90.20 & 15.79\% & 28.44 & 82.15 & 27.22\% & 18.74 & 37.62 & 14.72\% & 27.52 & 179.88 & 34.02\% \\
     & RPMixer & 17.50 & 28.90 & 13.03\% & 23.99 & 40.83 & 20.20\% & 18.59 & 30.47 & 14.45\% & 23.43 & 38.92  & 19.23\% \\
     \cline{2-14}
     & TSMixer+EiM  & 18.27 & 30.92 & 13.31\% & 18.71 & 31.72 & 13.75\% & 18.69 & 31.51 & 13.98\% & 18.37 & 31.20  & 13.57\% \\
     & RPMixer+EiM  & 17.68 & 30.28 & 12.74\% & 17.64 & 30.29 & 12.59\% & 17.59 & 30.17 & 12.69\% & 17.99 & 30.68  & 13.11\% \\
     & iXFMR+EiM  & 17.16 &\textbf{ 29.67 }& 11.91\% & \textbf{17.08} & \textbf{29.63} & \textbf{11.89}\% & 17.68 & 30.45 & 12.78\% & \textbf{17.13} & \textbf{29.61}  & \textbf{12.02}\% \\
     \cline{2-14}
     & \name{} (Ours)  & \textbf{17.10} & 29.74 & \textbf{11.84}\% & 17.42 & 30.03 & 12.37\% & \textbf{17.16} & \textbf{29.66} & \textbf{12.04}\% & 17.38 & 30.04  & 12.25\% \\
     \shline
\end{tabular}%
}
\end{center}
\end{table*}

%% file: insert/fig_dataset.tex
\begin{figure}[t]
\centerline{
\includegraphics[width=0.8\linewidth]{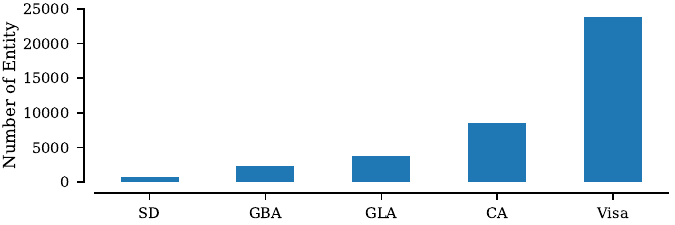}
}
\caption{Dataset statistics.}
\label{fig:dataset}
\end{figure}


%% file: insert/tab_transaction.tex
\begin{table*}[pt]
\caption{Performance comparisons on proprietary transaction volume dataset.
We bold the best-performing results.
The performance is reported at prediction steps 24 (Day 1), 72 (Day 3), and 168 (Day 7), as well as the average across all steps.}
\label{tab:transaction}
\begin{center}
\resizebox{0.75\linewidth}{!}{%
\begin{tabular}{lccc|ccc|ccc|ccc}
    \shline
    \multirow{2}{*}{Method} & \multicolumn{3}{c}{Horizon 1 Day} & \multicolumn{3}{c}{Horizon 3 Day} & \multicolumn{3}{c}{Horizon 7 Day} & \multicolumn{3}{c}{Average} \\ \cline{2-13} 
     & MAE & RMSE & MAPE & MAE & RMSE & MAPE & MAE & RMSE & MAPE & MAE & RMSE & MAPE \\ 
     \hline \hline
Linear  & 49.71 & 2187.15 & 70.00\% & 52.40 & 2299.22 & 73.52\% & 42.99 & 1685.41 & 71.39\% & 49.21 & 2088.29 & 71.52\% \\
TSMixer    & 49.43 & 2087.96 & 83.86\% & 48.15 & 1967.83 & 77.15\% & 44.96 & 1772.48 & 76.86\% & 46.11 & 1835.36 & 74.82\% \\
RPMixer    & 57.50 & 2490.48 & 93.37\% & 59.13 & 2552.62 & 95.02\% & 53.47 & 2195.45 & 92.64\% & 57.60 & 2540.66 & 93.98\% \\
\hline
TSMixer+EiM  & 39.63 & 1624.31 & 66.31\% & 43.31 & 1773.76 & 70.33\% & 39.68 & 1511.46 & 70.66\% & 42.06 & 1692.32 & 69.49\% \\
RPMixer+EiM & 37.76 & 1581.57 & 62.65\% & 39.73 & 1575.00 & 63.43\% & 37.45 & 1401.51 & 66.29\% & 39.16 & 1520.26 & 65.89\% \\
iXFMR+EiM  & 34.20 & 1257.40 & 62.54\% & 36.97 & 1381.20 & \textbf{62.38}\% & 36.82 & 1347.03 & 68.07\% & 35.93 & 1308.49 & 63.77\% \\
\hline
\name{} (Ours)  & \textbf{32.25} & \textbf{1146.07} & \textbf{59.06}\% & \textbf{36.39} & \textbf{1310.34} & 62.68\% & \textbf{34.50} & \textbf{1178.00} & \textbf{63.46}\% & \textbf{34.77} & \textbf{1230.72} & \textbf{63.49}\% \\
     \shline
\end{tabular}%
}
\end{center}
\end{table*}

%% file: insert/fig_qualitative.tex
\begin{figure}[htp]
\centerline{
\includegraphics[width=0.85\linewidth]{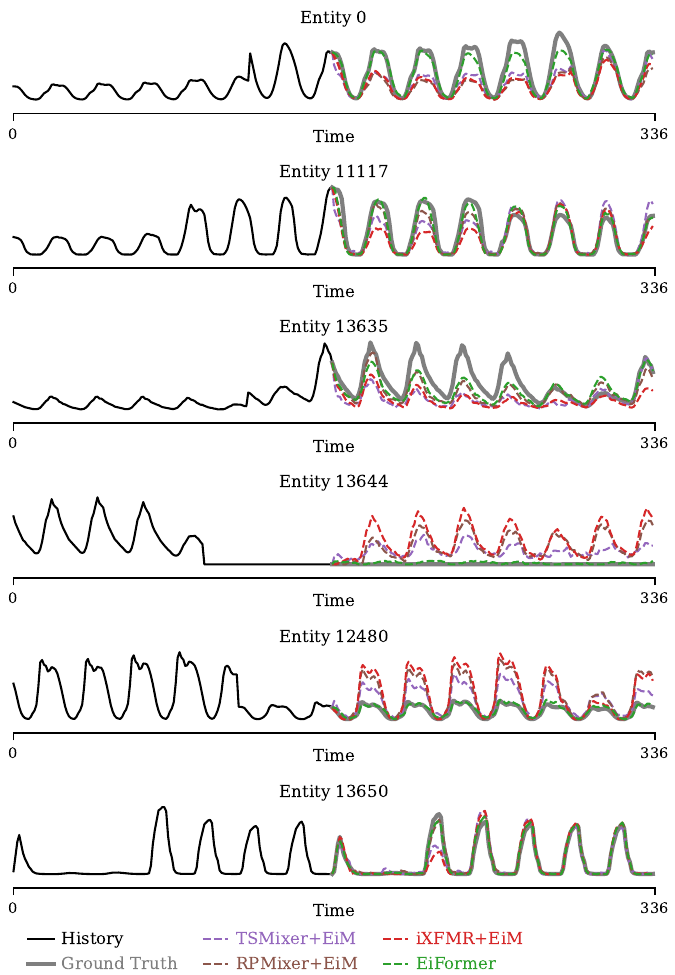}
}
\caption{
Qualitative comparison between the prediction generated by \name{} with alternatives for transaction dataset.
}
\label{fig:qualitative}
\end{figure}

%% file: insert/fig_net_analysis_linear_cka.tex
\begin{figure}[t]
    \begin{subfigure}{0.15\textwidth}
        \centering
        \includegraphics[width=\textwidth]{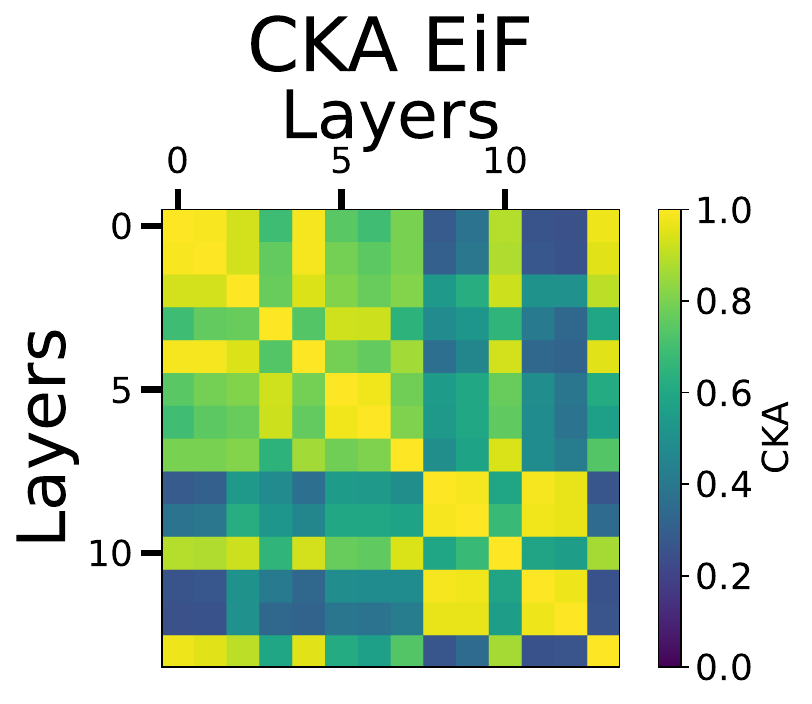}
        \caption{EiFormer}
        \label{fig:CKA_EiF}
    \end{subfigure}
    \begin{subfigure}{0.15\textwidth}
        \centering
        \includegraphics[width=\textwidth]{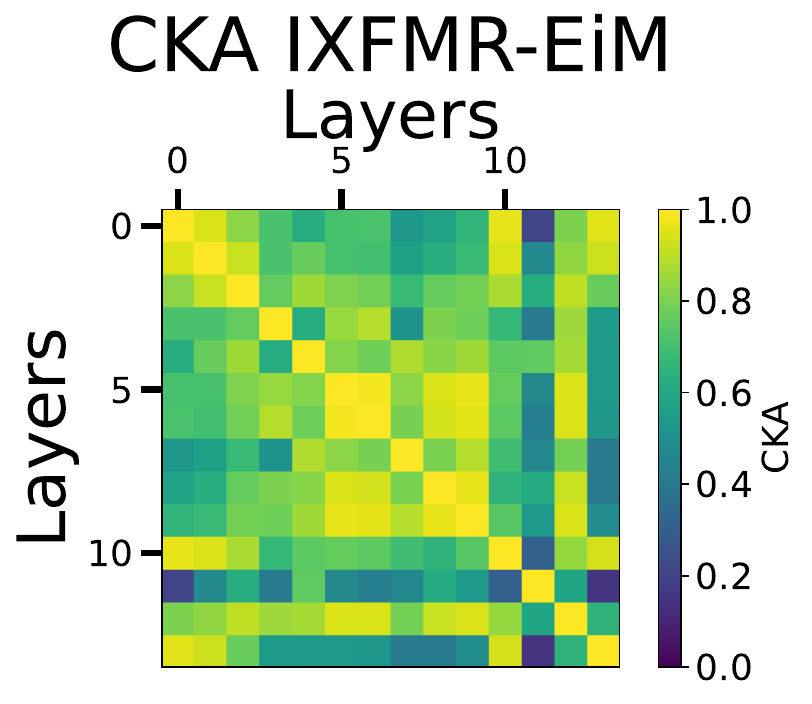}
        \caption{iXFMR+EiM}
        \label{fig:CKA_IXFMR-EiM}
    \end{subfigure}
    \begin{subfigure}{0.15\textwidth}
        \centering
        \includegraphics[width=\textwidth]{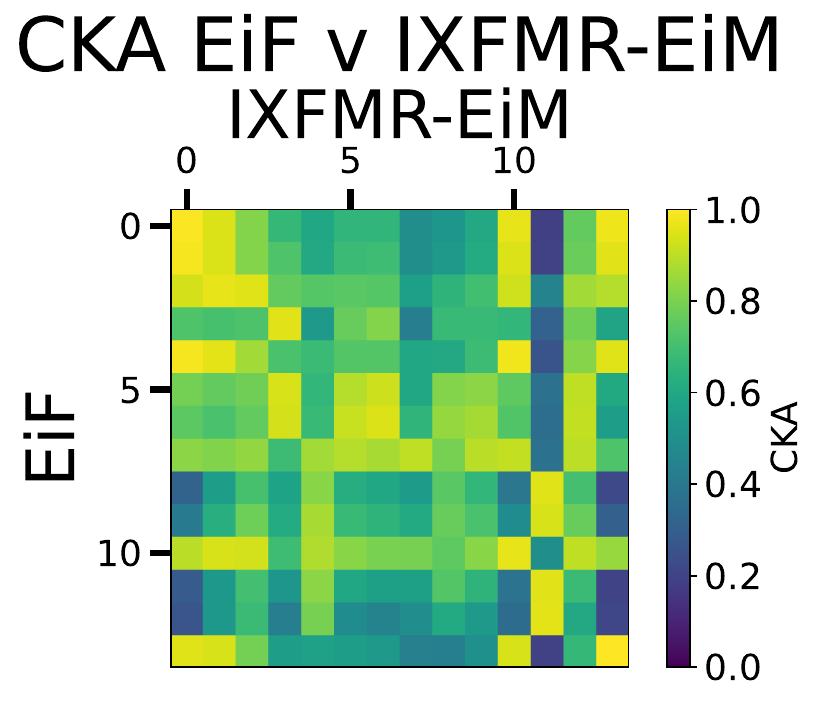}
        \caption{EiF v iXFMR+EiM}
        \label{fig:CKA_EiF_IXFMR-EiM}
    \end{subfigure}
    \caption{Linear CKA based analysis of EiFormer and iXMFR+EiM.}
    \label{fig:Linear_CKA_EiF_IXFMR-EiM}
\end{figure}

%% file: insert/fig_cost_forward.tex
\begin{figure}[t]
\centerline{
\includegraphics[width=0.8\linewidth]{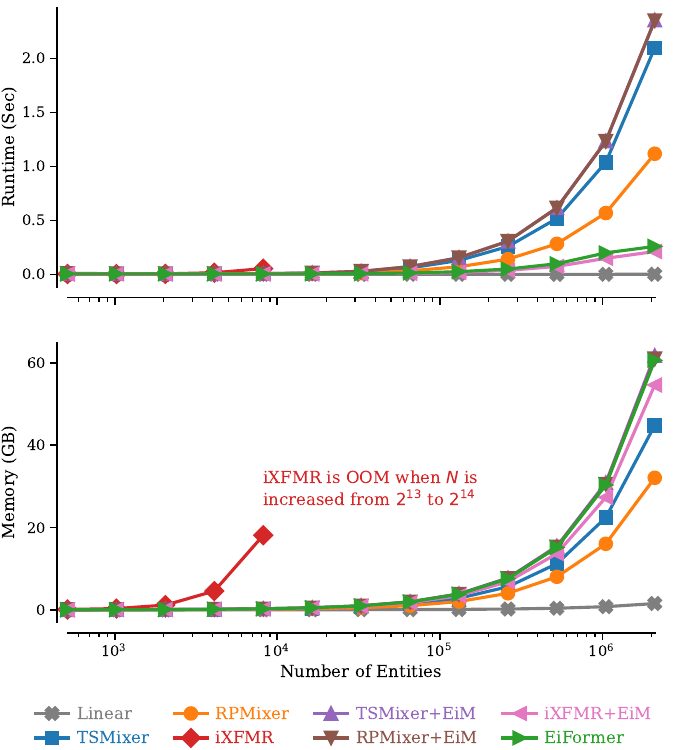}
}
\caption{
Computational cost associated with the forward pass for runtime and GPU memory usage.
}
\label{fig:cost_forward}
\end{figure}

%% file: insert/fig_param_lr.tex
\begin{figure}[t]
\centerline{
\includegraphics[width=0.85\linewidth]{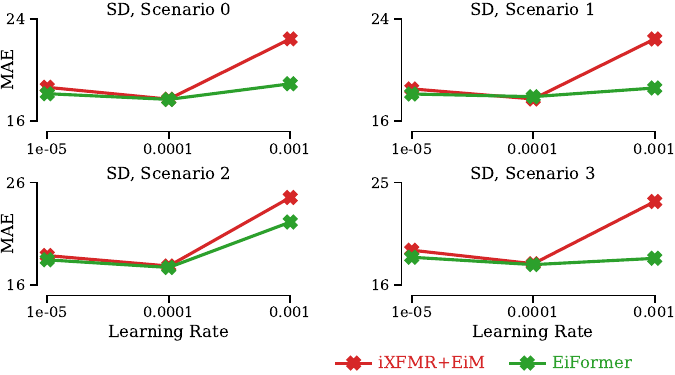}
}
\caption{
\textit{Learning rate} sensitivity analysis using MAE.
}
\label{fig:param_lr}
\end{figure}

%% file: insert/fig_param_layer.tex
\begin{figure}[t]
\centerline{
\includegraphics[width=0.85\linewidth]{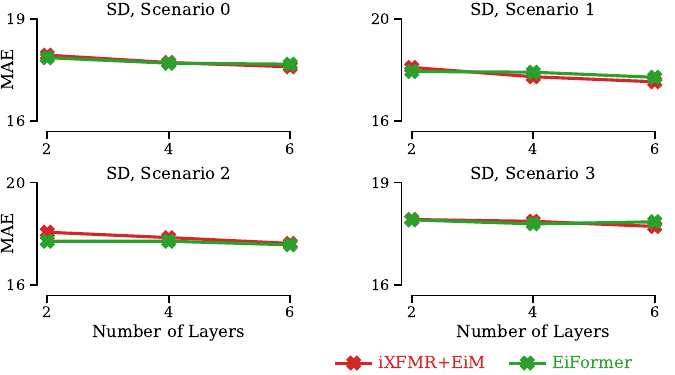}
}
\caption{
\textit{Number of layers} sensitivity analysis using MAE.
}
\label{fig:param_layer}
\end{figure}

%% file: insert/fig_param_neuron.tex
\begin{figure}[htp]
\centerline{
\includegraphics[width=0.85\linewidth]{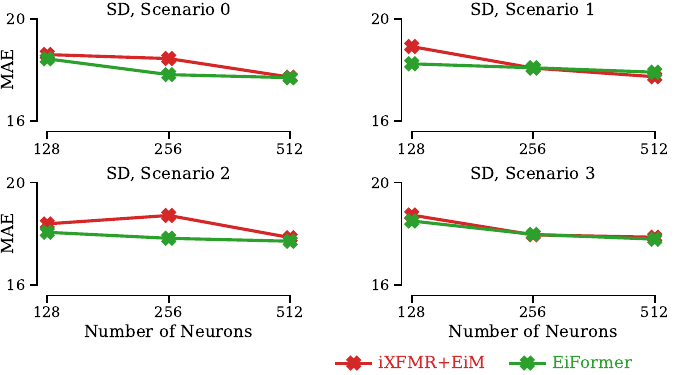}
}
\caption{
\textit{Number of neurons} sensitivity analysis using MAE.
}
\label{fig:param_neuron}
\end{figure}

%% file: insert/tab_largest_benchmark.tex
\begin{table*}[pt]
\caption{Performance comparisons. 
We bold the best-performing results.
The performance reported in the ``Average" column is computed by averaging over 12 predicted time steps.}
\label{tab:largest_benchmark}
\begin{center}
\resizebox*{!}{0.93\textheight}{%
\begin{tabular}{llccc|ccc|ccc|ccc}
    \shline
    \multirow{2}{*}{Data} & \multirow{2}{*}{Method} & \multicolumn{3}{c}{Horizon 3} & \multicolumn{3}{c}{Horizon 6} & \multicolumn{3}{c}{Horizon 12} & \multicolumn{3}{c}{Average} \\ \cline{3-14} 
     &  & MAE & RMSE & MAPE & MAE & RMSE & MAPE & MAE & RMSE & MAPE & MAE & RMSE & MAPE \\ 
     \hline \hline
     \multirow{20}{*}{SD} & HL  & 33.61 & 50.97 & 20.77\% & 57.80 & 84.92 & 37.73\% & 101.74 & 140.14 & 76.84\% & 60.79 & 87.40 & 41.88\% \\
     & 1NN & 21.79 & 35.15 & 13.79\% & 25.64 & 41.59 & 17.05\% & 30.77 & 50.59 & 22.38\% & 25.47 & 41.36 & 17.25\% \\
     & Linear & 20.58 & 33.30 & 12.98\% & 27.00 & 43.87 & 18.20\% & 32.35 & 53.51 & 22.38\% & 25.85 & 42.35 & 17.10\% \\
     & LSTM & 19.17 & 30.75 & 11.85\% & 26.11 & 41.28 & 16.53\% & 38.06 & 59.63 & 25.07\% & 26.73 & 42.14 & 17.17\% \\
     & ASTGCN & 20.09 & 32.13 & 13.61\% & 25.58 & 40.41 & 17.44\% & 32.86 & 52.05 & 26.00\% & 25.10 & 39.91 & 18.05\% \\
     & DCRNN & 17.01 & 27.33 & 10.96\% & 20.80 & 33.03 & 13.72\% & 26.77 & 42.49 & 18.57\% & 20.86 & 33.13 & 13.94\% \\
     & AGCRN & 16.05 & 28.78 & 11.74\% & 18.37 & 32.44 & 13.37\% & 22.12 & 40.37 & 16.63\% & 18.43 & 32.97 & 13.51\% \\
     & STGCN & 18.23 & 30.60 & 13.75\% & 20.34 & 34.42 & 15.10\% & 23.56 & 41.70 & 17.08\% & 20.35 & 34.70 & 15.13\% \\
     & GWNET  & 15.49 & 25.45 & 9.90\% & 18.17 & 30.16 & 11.98\% & 22.18 & 37.82 & 15.41\% & 18.12 & 30.21 & 12.08\% \\
     & STGODE  & 16.76 & 27.26 & 10.95\% & 19.79 & 32.91 & 13.18\% & 23.60 & 41.32 & 16.60\% & 19.52 & 32.76 & 13.22\% \\
     & DSTAGNN  & 17.83 & 28.60 & 11.08\% & 21.95 & 35.37 & 14.55\% & 26.83 & 46.39 & 19.62\% & 21.52 & 35.67 & 14.52\% \\
     & DGCRN  & 15.24 & 25.46 & 10.09\% & 17.66 & 29.65 & 11.77\% & 21.38 & 36.67 & 14.75\% & 17.65 & 29.70 & 11.89\% \\
     & D$^2$STGNN  & 14.85 & 24.95 & 9.91\% & 17.28 & 29.05 & 12.17\% & 21.59 & 35.55 & 16.88\% & 17.38 & 28.92 & 12.43\% \\
     \cline{2-14}
     & TSMixer  & 17.13 & 27.42 & 11.35\% & 19.30 & 31.07 & 12.50\% & 22.03 & 35.70 & 14.26\% & 19.06 & 30.66 & 12.55\% \\
     & RPMixer  & \textbf{15.12} & \textbf{24.83} & 9.97\% & \textbf{17.04} & \textbf{28.24} & \textbf{10.98}\% & \textbf{19.60} & \textbf{32.96} & 13.12\% & \textbf{16.90} & \textbf{27.97} & 11.07\% \\
     & iXFMR  & 15.24 & 25.33 & 9.59\%  & 17.52 & 29.37 & \textbf{10.98}\% & 20.46 & 34.74 & \textbf{12.98}\% & 17.34 & 29.06 & \textbf{10.93}\% \\
     \cline{2-14}
     & TSMixer+EiM  & 16.21 & 26.92 & 10.12\% & 19.02 & 32.16 & 12.13\% & 22.59 & 39.16 & 15.26\% & 18.89 & 31.89 & 12.24\% \\
     & RPMixer+EiM  & 15.75 & 26.22 & 9.82\%  & 18.34 & 30.89 & 11.54\% & 21.50 & 37.42 & 13.89\% & 18.13 & 30.63 & 11.54\% \\
     & iXFMR+EiM  & 15.34 & 25.83 & \textbf{9.31}\%  & 17.99 & 30.66 & 11.12\% & 21.03 & 36.69 & 13.46\% & 17.72 & 30.24 & 11.03\% \\
     \cline{2-14}
     & \name{} (Ours) & 15.28 & 25.72 & 9.39\%  & 17.81 & 30.32 & 11.08\% & 21.15 & 36.72 & 13.60\% & 17.70 & 30.10 & 11.09\% \\
     \hline \hline
     \multirow{19}{*}{GBA} & HL & 32.57 & 48.42 & 22.78\% & 53.79 & 77.08 & 43.01\% & 92.64 & 126.22 & 92.85\% & 56.44 & 79.82 & 48.87\% \\
     & 1NN & 24.84 & 41.30 & 17.70\% & 29.31 & 48.56 & 22.92\% & 35.22 & 58.44 & 31.07\% & 29.10 & 48.23 & 23.14\% \\
     & Linear & 21.55 & 34.79 & 17.94\% & 27.24 & 43.36 & 23.66\% & 31.50 & 51.56 & 26.18\% & 26.12 & 42.14 & 22.10\% \\
     & LSTM & 20.41 & 33.47 & 15.60\% & 27.50 & 43.64 & 23.25\% & 38.85 & 60.46 & 37.47\% & 27.88 & 44.23 & 24.31\% \\
     & ASTGCN & 21.40 & 33.61 & 17.65\% & 26.70 & 40.75 & 24.02\% & 33.64 & 51.21 & 31.15\% & 26.15 & 40.25 & 23.29\% \\
     & DCRNN & 18.25 & 29.73 & 14.37\% & 22.25 & 35.04 & 19.82\% & 28.68 & 44.39 & 28.69\% & 22.35 & 35.26 & 20.15\% \\
     & AGCRN& 18.11 & 30.19 & 13.64\% & 20.86 & 34.42 & 16.24\% & 24.06 & 39.47 & 19.29\% & 20.55 & 33.91 & 16.06\% \\
     & STGCN & 20.62 & 33.81 & 15.84\% & 23.19 & 37.96 & 18.09\% & 26.53 & 43.88 & 21.77\% & 23.03 & 37.82 & 18.20\% \\
     & GWNET & 17.74 & 28.92 & 14.37\% & 20.98 & 33.50 & 17.77\% & 25.39 & 40.30 & 22.99\% & 20.78 & 33.32 & 17.76\% \\
     & STGODE & 18.80 & 30.53 & 15.67\% & 22.19 & 35.91 & 18.54\% & 26.27 & 43.07 & 22.71\% & 21.86 & 35.57 & 18.33\% \\
     & DSTAGNN & 19.87 & 31.54 & 16.85\% & 23.89 & 38.11 & 19.53\% & 28.48 & 44.65 & 24.65\% & 23.39 & 37.07 & 19.58\% \\
     & DGCRN & 18.09 & 29.27 & 15.32\% & 21.18 & 33.78 & 18.59\% & 25.73 & 40.88 & 23.67\% & 21.10 & 33.76 & 18.58\% \\
     & D$^2$STGNN & 17.20 & \textbf{28.50} & 12.22\% & 20.80 & 33.53 & 15.32\% & 25.72 & 40.90 & 19.90\% & 20.71 & 33.44 & 15.23\% \\
     \cline{2-14}
     & TSMixer & 17.57 & 29.22 & 14.14\% & 19.85 & 32.64 & 16.95\% & 22.27 & 37.60 & 18.63\% & 19.58 & 32.56 & 16.58\% \\
     & RPMixer  & 17.35 & 28.69 & 13.42\% & 19.44 & \textbf{32.04} & 15.61\% & \textbf{21.65} & \textbf{36.20} & 17.42\% & 19.06 & \textbf{31.54} & 15.09\% \\
     \cline{2-14}
     & TSMixer+EiM  & 17.33 & 29.23 & 13.52\% & 20.04 & 33.76 & 16.13\% & 22.93 & 38.85 & 18.13\% & 19.68 & 33.20 & 15.65\% \\
     & RPMixer+EiM  & 16.73 & 28.70 & 12.38\% & 19.34 & 33.07 & 15.02\% & 22.09 & 38.37 & 17.64\% & 18.98 & 32.62 & 14.58\% \\
     & iXFMR+EiM   & \textbf{16.47} & 28.60 & 11.74\% & \textbf{18.99} & 32.92 & \textbf{13.70}\% & 21.78 & 38.41 & \textbf{16.19}\% & \textbf{18.72} & 32.58 & \textbf{13.55}\% \\
     \cline{2-14}
     & \name{} (Ours)& 16.49 & 28.58 & \textbf{11.58}\% & 19.11 & 33.05 & 14.00\% & 21.79 & 38.26 & 16.21\% & 18.76 & 32.59 & 13.65\% \\
     \hline \hline
     \multirow{17}{*}{GLA} & HL & 33.66 & 50.91 & 19.16\% & 56.88 & 83.54 & 34.85\% & 98.45 & 137.52 & 71.14\% & 59.58 & 86.19 & 38.76\% \\
     & 1NN &  23.23 & 38.69 & 13.44\% & 27.75 & 45.92 & 17.07\% & 33.49 & 55.51 & 22.86\% & 27.49 & 45.57 & 17.28\% \\
     & Linear & 21.32 & 34.48 & 13.35\% & 27.45 & 43.83 & 17.79\% & 32.50 & 52.69 & 21.76\% & 26.40 & 42.56 & 17.16\% \\
     & LSTM &  20.09 & 32.41 & 11.82\% & 27.80 & 44.10 & 16.52\% & 39.61 & 61.57 & 25.63\% & 28.12 & 44.40 & 17.31\% \\
     & ASTGCN & 21.11 & 34.04 & 12.29\% & 28.65 & 44.67 & 17.79\% & 39.39 & 59.31 & 28.03\% & 28.44 & 44.13 & 18.62\% \\
     & DCRNN & 18.33 & 29.13 & 10.78\% & 22.70 & 35.55 & 13.74\% & 29.45 & 45.88 & 18.87\% & 22.73 & 35.65 & 13.97\% \\
     & AGCRN & 17.57 & 30.83 & 10.86\% & 20.79 & 36.09 & 13.11\% & 25.01 & 44.82 & 16.11\% & 20.61 & 36.23 & 12.99\% \\
     & STGCN & 19.87 & 34.01 & 12.58\% & 22.54 & 38.57 & 13.94\% & 26.48 & 45.61 & 16.92\% & 22.48 & 38.55 & 14.15\% \\
     & GWNET & 17.30 & 27.72 & 10.69\% & 21.22 & 33.64 & 13.48\% & 27.25 & 43.03 & 18.49\% & 21.23 & 33.68 & 13.72\% \\
     & STGODE  & 18.46 & 30.05 & 11.94\% & 22.24 & 36.68 & 14.67\% & 27.14 & 45.38 & 19.12\% & 22.02 & 36.34 & 14.93\% \\
     & DSTAGNN & 19.35 & 30.55 & 11.33\% & 24.22 & 38.19 & 15.90\% & 30.32 & 48.37 & 23.51\% & 23.87 & 37.88 & 15.36\% \\
     \cline{2-14}
     & TSMixer  & 20.38 & 224.82 & 13.62\% & 22.90 & 229.86 & 15.51\% & 23.63 & 135.09 & 15.56\% & 22.12 & 207.68 & 14.87\% \\
     & RPMixer & 16.49 & 26.75 & 9.75\% & 18.82 &\textbf{ 30.56} & 11.58\% & \textbf{21.18} & \textbf{35.10 }& 13.46\% & 18.46 & \textbf{30.13} & 11.34\% \\
     \cline{2-14}
     & TSMixer+EiM & 17.18 & 28.34 & 10.52\% & 20.07 & 33.41 & 12.41\% & 23.24 & 39.38 & 14.98\% & 19.76 & 32.95 & 12.46\% \\
     & RPMixer+EiM & 16.04 & 26.88 & 9.39\%  & 18.86 & 31.62 & 11.28\% & 22.17 & 38.00 & 13.76\% & 18.60 & 31.40 & 11.22\% \\
     & iXFMR+EiM  & \textbf{15.71} & \textbf{26.52} & \textbf{8.97}\%  & \textbf{18.38} & 31.13 & \textbf{10.84}\% & 21.50 & 37.19 & \textbf{13.17}\% & \textbf{18.14} & 30.88 & \textbf{10.76}\% \\
     \cline{2-14}
     & \name{} (Ours)  & 15.75 & 26.67 & 8.99\%  & 18.51 & 31.55 & 10.92\% & 21.59 & 37.55 & 13.27\% & 18.22 & 31.17 & 10.80\% \\
     \hline \hline
     \multirow{14}{*}{CA} & HL  & 30.72 & 46.96 & 20.43\% & 51.56 & 76.48 & 37.22\% & 89.31 & 125.71 & 76.80\% & 54.10 & 78.97 & 41.61\% \\
     & 1NN & 21.88 & 36.67 & 15.19\% & 25.94 & 43.36 & 19.19\% & 31.29 & 52.52 & 25.65\% & 25.76 & 43.10 & 19.43\% \\
     & Linear & 19.82 & 32.39 & 14.73\% & 25.20 & 40.97 & 19.24\% & 29.80 & 49.34 & 23.09\% & 24.32 & 39.88 & 18.52\% \\
     & LSTM & 19.01 & 31.21 & 13.57\% & 26.49 & 42.54 & 20.62\% & 38.41 & 60.42 & 31.03\% & 26.95 & 43.07 & 21.18\% \\
     & DCRNN  & 17.52 & 28.18 & 12.55\% & 21.72 & 34.19 & 16.56\% & 28.45 & 44.23 & 23.57\% & 21.81 & 34.35 & 16.92\% \\
     & STGCN  & 19.14 & 32.64 & 14.23\% & 21.65 & 36.94 & 16.09\% & 24.86 & 42.61 & 19.14\% & 21.48 & 36.69 & 16.16\% \\
     & GWNET & 16.93 & 27.53 & 13.14\% & 21.08 & 33.52 & 16.73\% & 27.37 & 42.65 & 22.50\% & 21.08 & 33.43 & 16.86\% \\
     & STGODE & 17.59 & 31.04 & 13.28\% & 20.92 & 36.65 & 16.23\% & 25.34 & 45.10 & 20.56\% & 20.72 & 36.65 & 16.19\% \\
     \cline{2-14}
     & TSMixer  & 18.40 & 106.28 & 14.30\% & 19.77 & 73.98 & 15.30\% & 22.56 & 87.56 & 17.80\% & 19.86 & 90.20 & 15.79\% \\
     & RPMixer  & 15.90 & 26.08 & 11.69\% & 17.79 & \textbf{29.37} & 13.23\% & \textbf{19.93} & \textbf{33.18} & 15.11\% & 17.50 & \textbf{28.90} & 13.03\% \\
     \cline{2-14}
     & TSMixer+EiM & 15.72 & 26.61 & 11.22\% & 18.62 & 31.32 & 13.42\% & 21.67 & 37.08 & 16.12\% & 18.27 & 30.92 & 13.31\% \\
     & RPMixer+EiM & 15.32 & 26.12 & 10.66\% & 17.97 & 30.65 & 12.90\% & 20.89 & 36.31 & 15.38\% & 17.68 & 30.28 & 12.74\% \\
     & iXFMR+EiM & 14.91 & 25.66 & 10.11\% & 17.41 & 30.00 & 12.11\% & 20.27 & 35.46 & 14.41\% & 17.16 & 29.67 & 11.91\% \\
     \cline{2-14}
     & \name{} (Ours) & \textbf{14.82} & \textbf{25.57} & \textbf{10.01}\% & \textbf{17.36} & 30.07 & \textbf{11.95}\% & 20.21 & 35.66 & \textbf{14.36}\% & \textbf{17.10} & 29.74 & \textbf{11.84}\% \\
     \shline
\end{tabular}%
}
\end{center}
\end{table*}

%% file: section/conclusion.tex
\section{Conclusion}

In this paper, we propose \name{}, an improved iTransformer model that addresses key challenges in forecasting large-scale spatial-temporal data.
By reducing time and space complexity from quadratic to linear via the proposed efficient latent attention, \name{} significantly enhances scalability while maintaining the ability to handle dynamic entity changes. The integration of a random projection mechanism further improves its predictive accuracy. 
Experiments on both public and proprietary datasets demonstrate \name{}'s superior performance in both scalability and predictive capabilities compared to baselines, showing the effectiveness of our method. 
